# A maximal-information color to gray conversion method for document images: Toward an optimal grayscale representation for document image binarization

Reza Farrahi Moghaddam, Shaohua Chen, Rachid Hedjam and Mohamed Cheriet

*Synchromedia Laboratory for Multimedia Communication in Telepresence,*
*École de technologie supérieure, Montreal (Quebec), Canada H3C 1K3*
*Tel.: +1-514-396-8972*
*Fax: +1-514-396-8595*
*imriss@ieee.org, rfarrahi@synchromedia.ca, mohamed.cheriet@etsmtl.ca*

**Abstract**

A novel method to convert color/multi-spectral images to gray-level images is introduced to increase the performance of document binarization methods. The method uses the distribution of the pixel data of the input document image in a color space to find a transformation, called *the dual transform*, which balances the amount of information on all color channels. Furthermore, in order to reduce the intensity variations on the gray output, a color reduction preprocessing step is applied. Then, a channel is selected as the gray value representation of the document image based on the homogeneity criterion on the text regions. In this way, the proposed method can provide a luminance-independent contrast enhancement. The performance of the method is evaluated against various images from two databases, the ICDAR'03 Robust Reading, the KAIST and the DIBCO'09 datasets, subjectively and objec-tively with promising results. The ground truth images for the images from the ICDAR'03 Robust Reading dataset have been created manually by the authors.

---

# 1 Introduction

The preservation of historical documents is a new cultural trend in almost every country in the world [1, 2, 19, 15]. High-quality images of the documents are being produced using sophisticated camera imaging systems, and are stored in various datasets across the globe. In historical and old documents, non-black colors have usually been used to differentiate between the section titles or important phrases from the main text, and also to mark the layout. Processing and understanding these document images is a difficult task facing the document image analysis and retrieval (DIAR) community. At the same time, camera-based images which capture readable human-generated artifacts, such as signs and documents, are becoming popular, and require to be processed. In this work, for the sake of simplicity, we consider any image containing readable human-generated artifacts as a document image. In many cases, a grayscale representation of a document image is preferred to be fed to the processing work flows, which use methods from various paradigms, such as statistical methods [30, 4] and variational methods [6, 8]. This is mainly because of the limitations of these document processing techniques or in order to reduce the computational cost. One of the most important processing steps in document image processing is binarization [12, 31, 27, 22]. Many state-of-the-art binarization methods require a *grayscale* representation of the input color image to proceed. Therefore, without a powerful conversion to grayscale that is resilience to degradation introduced because of color conversion, these methods cannot achieve their highest capability. This is mainly because of lack of correlation between the variation of color on a document image and its text; the color data on a document images usually carry the layout data, and has less relation to its actual content. Although the conversion of a color image to grayscale is usually considered equivalent to calculating *luminance*, its impact on the subsequent processes, such as binarization, cannot be ignored. In this work, we provide a generic and information-based conversion to grayscale in order to improve the performance of gray-based binarization methods, and therefore we do not consider binarization methods which directly work on the color images.

Traditionally, as mentioned above, the luminance map of a color image is considered as its grayscale representation. The benefit of this definition, which also implicitly defines luminance, is that it tries to keep as much color data as possible on the gray image, thereby enabling succeeding processing steps to differentiate between regions with different colors on the original image. For example, in Figure 1, the luminance map provides a color-sensitive grayscale representation of the color image shown. However, depending on the target application, this may not always be the objective [25]. For example, in [13], a transformation based on modeling figure–ground segregation has been proposed to reflect the brightness perception on the output image. A brief discussion on some of related work is presented in section 3. For our application, with the goal of binarizing document images or those images with a large amount of structural information, lower sensitivity to the color values of the objects is preferred. The output of our proposed method, which will be called

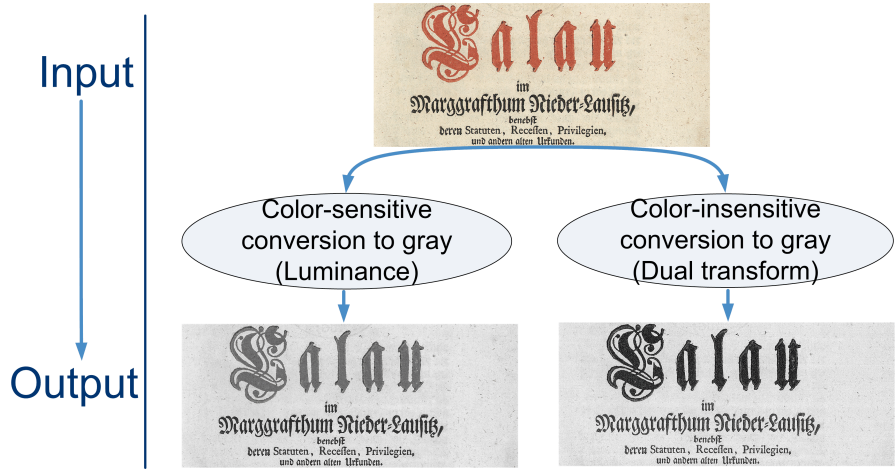

Fig. 1. A schematic representation of the difference between color-sensitive and color-insensitive conversions of color images to grayscale.

*the dual transform*, is also shown in Figure 1. We call this transform *dual* as it is in contrast to the Principal Component Analysis (PCA) method that tries to compress as much as possible information in one channel.

In our work here, a novel method of conversion to gray is introduced, which is insensitive to color data. This conversion method, the dual transform, balances the input image data among all color channels. Using a color reduction process, possible amplified color variations are avoided. This process also uses an interpolation step to avoid introducing artificial color jumps on the image. Finally, one of the color channels is selected as the output gray image based on its variation on the text regions. Some samples of the document images used in the work are shown in Figure 2.

The paper is organized as follows. The problem statement is presented in section 2. In section 3, a few conversion methods are discussed. The dual transform is introduced in section 4. Section 5 provides the details of the color reduction and interpolation step. The experimental results and discussions are presented in section 6. Finally, the conclusion and some prospects for future work are discussed in section 7.

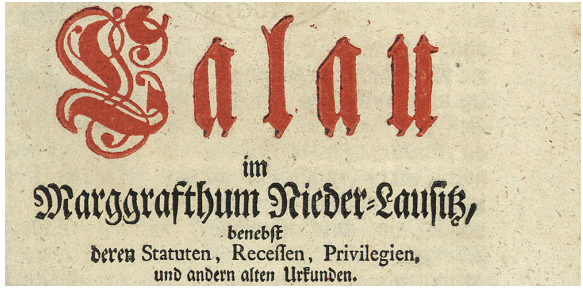 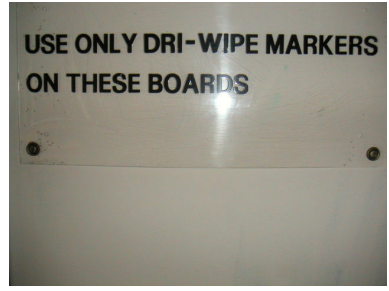

(a) (b)

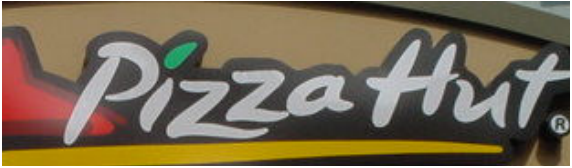 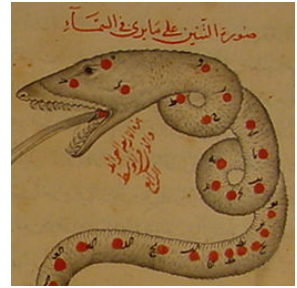

(c) (d)

Fig. 2. Some sample images. a) An image from the DIBCO'09 dataset. b) An image from the ICDAR'03 Robust Reading dataset. c) An image from the KAIST dataset. d) An image from the Juma Al Majid manuscripts. For more details, please see section 6.

## 2 Problem Statement

A color/multi-spectral document image $u$ is given, $u : \Omega \to \mathbb{R}^{n_c}$, where $\Omega \subset \mathbb{R}^2$ is an open rectangle and $n_c$ is the number of color channels. For RGB color images, $n_c = 3$. A single-value equivalent of $u$, called the gray-level image and denoted $I$, is required, $I : \Omega \to \mathbb{R}$. The goal is that $I$ contain as much of the structural information of the input image as possible and is color-insensitive. It is assumed that the color space is represented by a set of unit vectors as its basis vectors. In order to find $I$, a linear transformation in the color space is considered, which is denoted $\tilde{T}$. The projection of the transformed color space onto one of its unit vectors is denoted $I$. This projection is based on an information-related optimization criterion.

# 3 Related Work

In this section, some related methods are described. They are arranged in two main categories: 1) sample-based methods; and 2) non sample-based methods.

## 3.1 Sample-based (supervised) methods

The conversion-to-gray problem can be seen as a dimension-reduction or feature-extraction one. Sample-based, or supervised, methods use the labels provided on the sample images to learn and label other images. They are usually referred to as the discriminant feature extraction techniques [11]. A famous example of these techniques is linear discriminant analysis (LDA) [9]. In a generalization, an analytical information-theoretical measure was introduced in [20], called information discriminant analysis (IDA) [5]. Sample-based methods outperform the others, but in many cases the samples are not available, or are not numerous enough, to infer the model. Also, they are more likely to be sensitive in the case of images that are out of the learned set. For these cases, non sample-based methods are used.

## 3.2 Non sample-based (unsupervised) methods

This category can be divided into two subcategories, depending on whether or not the method uses the image data.

### 3.2.1 Non adaptive methods

As mentioned in the Introduction, the *luminance* of the color image is usually used as one of the possible conversions to gray (the *luminance* method). By definition[1], the luminance-based gray value of a pixel on the image domain is calculated as follows:

$$I_{\text{lum}}(x) = 1 - \text{LUM}(x) = 1 - \{0.27 u_{\text{red}}(x) + 0.67 u_{\text{green}}(x) + 0.06 u_{\text{blue}}(x)\} \quad (1)$$

where $I_{\text{lum}}(x)$ and $\text{LUM}(x)$ are the calculated gray value and luminance at pixel $x$ respectively. The gray values are calculated in the BW10 representation [7], in which zero values are shown by white pixels, black pixels corresponding to values equal to 1. Also, for example, $u_{\text{red}}$ is the red channel of the input color image. In an illustrative visualization, the isogray curves of the luminance conversion on a two-dimensional subspace of the color space (only red and green colors) are shown in Figure 3(a). As can be seen from the figure, the conversion behaves differently for different color channels. This is a preferred behavior when we want to transfer as much as possible color data to the grayscale representation. However, as discussed in the introduction, we need a color-insensitive conversion for binarization of multicolor document images.

Another conversion, which is more popular in the case of document image processing, is *average* conversion. In this technique, the gray value at a pixel is the mean value of all the channels:

$$I_{\text{avg}}(x) = 1 - \frac{1}{n_c} \sum_{i=1}^{n_c} u_i(x) \qquad (2)$$

where $I_{\text{avg}}$ is the gray value image calculated based on the average conversion method. The isogray curves of this method are shown in Figure 3(b). This method is still sensitive to colors but shows a symmetrical behavior that is more preferred in the case of document images which do not have any preference for a specific color.

---

[1] As defined by ITU-R Recommendation BT.601.

As discussed, less color sensitivity is of great interest in document image processing, especially in the case of historical document images that are written in many colors. In historical and old documents, non-black colors, such as red color, have been used to specify the layout and sections. However, because of aging and other deteriorations, the colored text is usually faded, and its luminance or average grayscale representations is very weak. In [7], a color-insensitive method has been introduced with promising results. We call it the *minimum average* method, as it uses the minimum channel to calculate the gray value [7]:

$$I_{\text{min-avg}}(x) = 1 - 0.5 \left( \frac{1}{n_c} \sum_{i=1}^{n_c} u_i(x) + \min_{i=1,\cdots,n_c} u_i(x) \right) \qquad (3)$$

where $I_{\text{min-avg}}(x)$ is the gray value calculated at pixel $x$. The isogray curves of this transformation, displayed in Figure 3(c), show less sensitivity with respect to color, i.e., the distances of the points on each isogray from the top-right corner (the white color) do not show a lot of variations. For an ideal conversion in this class, the distance of an isogray to the white color should be a constant in order to achieve complete symmetry with respect to all the colors. Therefore, we can propose that the *ideal* method in this category should yield completely circular curves on this subspace (as shown in Figure 3(d)), which would provide a constant distance to the white color. However, our focus here is on an adaptive solution, and therefore this direction will not be investigated. The adaptive methods are discussed in the next subsection.

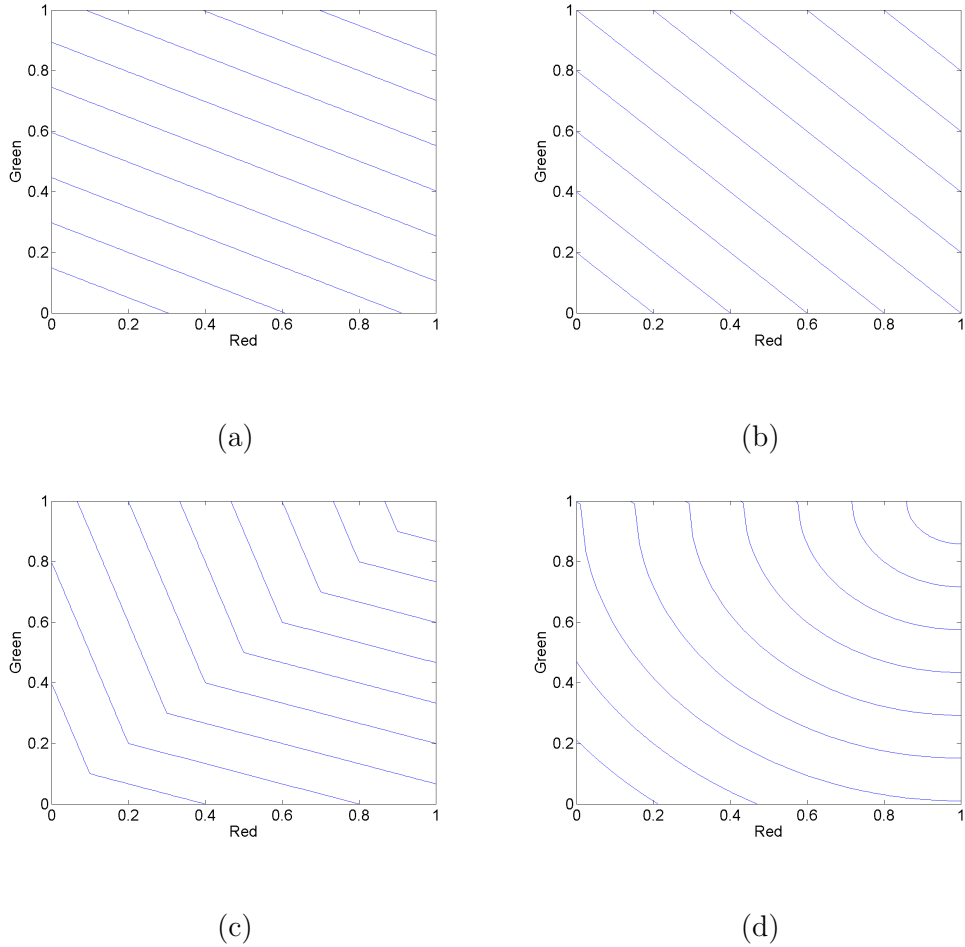

Fig. 3. A 2-dimensional subspace (red and green) of the color space illustration of the gray converters. a) The luminance method. b) The average method. c) The min-average method. d) An ideal converter. The blue lines are isogray curves corresponding to different gray levels.

*3.2.2 Adaptive methods*

Statistical methods, such as Principal Component Analysis (PCA) [23, 10] and independent component analysis (ICA) have been used for feature extraction purposes in the case of color/multi-spectral or double-sided document images [29, 28, 26]. The availability of various channels in these images enables these methods to differentiate between different sources of information on the input color image. Although this ability is of great interest, it runs counter to the goal of having as much of the information as possible on just one output channel (the gray-level output). However, PCA is able to keep most of the significant data on one channel. Therefore, we consider this method in our experiments. PCA consists of a transformation of the color space, the principal axes of which are along the most variable direction. The unit vectors of the axes on the new coordinate system after the PCA transformation are denoted $\{p_i, i = 1, \cdots, n_c\}$. The unit vectors make up an orthogonal system, the corresponding eigenvalues of which are $\{\lambda_i, i = 1, \cdots, n_c\}$, where $\lambda_1$ is the largest eigenvalue. In the following section, the proposed dual transform is introduced, its formulation described using the PCA notation.

## 4 The Dual Space Conversion

For a better understanding of the dual transform, we start with the PCA transformation. it is worth noting that we calculate a unique PCA transformation for each input image based on the distribution of its pixels in the color space. It is assumed that the pixels of the input color image implicitly form two classes in the color space. Each of these classes consists of many sub classes in the color space. Although the PCA provides the maximum distance between the two main classes, it also introduces big gaps between the sub classes. In other words, an optimal criterion is required which balances the inter- and intra- distances of the classes prior to their classification. Here, we use the criterion of maximal balance of the information (energy) distribution among all channels of the color/multi-spectral space. In an abstract way, this criterion can be expressed as follows:

$$\tilde{T} = \arg\max_{T} \; \{\Pi_{i=1}^{n_c} E_i(T(u))\} \tag{4}$$

where $T$ can be any transformation in the color space, $n_c$ is the number of colors/channels, and $E_i$ is a measure, for example the entropy of the concurrence matrix, of the information of the $i^{th}$ channel. The above optimization problem finds an optimal transformation, after which the amount of information on each channel is almost the same as on the others. As the total amount of information on an image is constant, and also $E_i \geq 0$, the mutual information differences between channels are not considered in the problem (4). Here, instead of direct optimization of the information measure, an equivalent problem is considered. It can be argued that the maximum amount of the image information is located along the principal component of the image pixels in the color space. Therefore, if a transformation symmetrically places its axes around the principal component, it will have almost equal shares of information on all its components. This can be performed by optimization of the following cost function:

$$\Pi_{i=1}^{n_c} \cos(e_i, p_1) \tag{5}$$

where $\{e_i, i = 1, \cdots, n_c\}$ is the set of the unit vectors of the transformation $T$, and $p_1$ is the principal component of PCA (with the highest eigenvalue). The eigenvalues of the PCA transformation are denoted by $\lambda_j$ and are sorted in descending order. Therefore, $\lambda_1$ is the largest eigenvalue and is associated with the principal component $p_1$. In practice, the distribution of the image pixels in the color space is not symmetric, and some part of the information is aligned with other components: $p_k$, $k = 2, \cdots, n_c$. In this case, the cost function should consider all the PCA components, but weight them according to their importance which can be measured based on their associated eigenvalues:

$$\sum_{k=1}^{n_c} \frac{\lambda_k}{\lambda_1} \Pi_{i=1}^{n_c} \cos\left(e_i, p_k\right) \tag{6}$$

Again, the mutual differences are not considered as all the terms are positive, and all the vectors are normalized. Therefore, the problem (4) can be replaced by a new problem that attempts to evenly place the distribution of pixels in the color space with respect to the axis of the target transformation. In this way, we can conclude that the share of energy in each channel is the same. The new problem can be written as follows:

$$\tilde{T} = \underset{T:\ ||e_i||=1,\ e_i \cdot e_j = \delta_{ij}}{\arg\max} \sum_{k=1}^{n_c} \frac{\lambda_k}{\lambda_1} \Pi_{i=1}^{n_c} \cos\left(e_i, p_k\right) \tag{7}$$

where $\{e_i, i = 1, \cdots, n_c\}$ is the set of the unit vectors of the transformation $T$, and $p_k$ is the $k^{th}$ unit vector of the color space after the PCA transformation. For the sake of simplicity, we assume that the input image is represented in the color space associated with the PCA transformation, i.e. the space resulting after a rotation defined by the PCA transformation. In this space: $\hat{k} = p_k$, where $\hat{k}$ is the unit vector along the $k^{th}$ axis. In this way, the $\cos(\cdot)$ functions can be replaced by the vectors projections:

$$\tilde{T} = \underset{T:\ ||e_i||=1,\ e_i \cdot e_j = \delta_{ij}}{\arg\max} \sum_{k=1}^{n_c} \frac{\lambda_k}{\lambda_1} \Pi_{i=1}^{n_c} e_{i,k}$$

where $e_{i,k}$ is the $k^{th}$ element of the $i^{th}$ unit vector. Each term in this cost function has a nonlinearity degree of the order of $n_c$. Another way to repre-sent the problem is to use just the differences. In this way, by minimizing the differences between the projections $e_{i,k}$, instead of maximizing their product, we can achieve the same symmetrical distribution of pixels around the principal components in the color space. This will reduce the nonlinearity of the cost function from an order of $n_c$ to quadratic because of using terms like $(e_{i,k} - e_{j,k})^2$ in the cost function. Now, the problem (4) can be replaced by a simpler form:

$$\tilde{T} = \operatorname*{arg\,min}_{T,\ ||e_i||=1,\ e_i \cdot e_j = \delta_{ij}} \sum_{k=1}^{n_c} \frac{\lambda_k}{\lambda_1} \left( \sum_{i=1}^{n_c-1} (e_{i,k} - e_{i+1,k})^2 \right) \qquad (8)$$

where $e_{i,k}$ is the $k^{th}$ element of the $i^{th}$ unit vector. As can be seen from the problem (8), not all the differences are required, and only $n_c - 1$ differences are sufficient to ensure equality of the projections. We chose to use sequential differences of the form $(e_{i,k} - e_{i+1,k})^2$ for the sake of simplicity. For the case $n_c = 3$, the problem (8) can be expanded as follows:

$$\begin{aligned}
\tilde{T} = \operatorname*{arg\,min}_{T:\ ||e_i||=1,\ e_i \cdot e_j = \delta_{ij}} & \left\{ \left( (e_{1,1} - e_{2,1})^2 + (e_{2,1} - e_{3,1})^2 \right) \right. \\
& + \frac{\lambda_2}{\lambda_1} \left( (e_{1,2} - e_{2,2})^2 + (e_{2,2} - e_{3,2})^2 \right) \\
& \left. + \frac{\lambda_3}{\lambda_1} \left( (e_{1,3} - e_{2,3})^2 + (e_{2,3} - e_{3,3})^2 \right) \right\}
\end{aligned} \qquad (9)$$

The problem (8), or the problem (9) in the case of $n_c = 3$, is the main definition of the dual transform. This nonlinear, constrained optimization problem can be solved using a numerical solver. We use a recursive quadratic programming (RQP) method as the numerical solver [24]. Having the unit vectors, $e_i$, $i = 1, \cdots, 3$, the color image is available in the dual space. Therefore, one of its channels can be used as the gray-level image. In the case of symmetrical distributions, any of the channels can play the same role as the gray-level image. However, for the asymmetrical distributions, one of the channels will have less sensitivity than the others. Below, a process to identify that channel is described.

Assume that $\tilde{u}$ is the new representation of $u$ after the dual transform. We want to select one of its channels, $\tilde{u}_{\tilde{i}}$, which has the lowest level of variation on the text regions. Let $u_{BW}$ be the ground-truth binarization of $u$ in the BW10 representation [7]. In practice, the ground truth is not available, and a rough binarization of $u$ can be used. We use Otsu's method applied on the minimum-average gray value equivalent of the input image. Then, we have the following optimization problem to find the index $\tilde{i}$ associated with the $\tilde{u}_{\tilde{i}}$:

$$\tilde{i} = \underset{i,\ i=1,\cdots,n_c}{\arg\min}\ \sigma_{\tilde{u}_i^{(t)}} \qquad (10)$$

where $\tilde{u}_i^{(t)}$ only contains the text regions of $\tilde{u}_i$ masked based on $u_{BW}$, and $\sigma$ is the standard deviation. The $\tilde{i}^{th}$ channel is less variable on the text regions, and therefore complies with the goal of having minimum sensitivity to the color data on the input image. In order to obtain the highest contrast, $\tilde{u}_{\tilde{i}}$ is rescaled to the full range of gray values. The final image is considered as the output of the dual transform, i.e. the gray value conversion of the input image, denoted $I_{\text{dual}}$. It is worth noting that if we choose the channel along $p_1$, the principal component, the variations will be the highest.

Also, it is worth mentioning that the proposed method may amplify interfering text patterns, such as the bleed-through effect. Although this will result in appearance of both text and interfering patterns on the final binarization, the interfering strokes can easily be removed using a color clustering process on the binarized regions. In contrast, the process of recovering missed text regions which suffer from faded ink or are in different colors is much more complicated. Therefore, the proposed method ensures that these regions are enhanced in terms of contrast before binarization.

## 5 Color Reduction and Interpolation

The dual transform can be independent of luminance. Therefore, it is possible that the variations on some regions may be amplified after the transformation. These amplified variations can show up on the final output. In order to avoid this phenomenon, a color reduction and interpolation step for document images is considered before the dual transform is applied. It is worth noting the color reduction process is not the main concern of this work, and comparison with state of the art, such as [21], [2] will be considered in future.

### 5.1 Color reduction using primary colors for document images

Some of color data on the image pixels have been modified by unwanted sources (noise) that should be excluded, and some are too few in number and can be easily discarded. The process of eliminating or discarding these data can be performed by dropping the corresponding color values from the possible color set, and recoloring the pixels with the remaining colors in the set. This process results in a more stable binarization without introducing perceptually-visible degradations.

Two questions need to be answered in order to perform the color reduction process: 1) how many color points should be used for image representation? and 2) what are those colors? The answer to the first question is the use of a histogram-based color reduction, and to the second is the construction of an indexed version of the image in order to represent it with only its primary colors.

---

[2] http://www.papamarkos.gr/index/category/134/

In this work, the CIELCh color space is used following the CIE standard [3]. This color space uses color data in the CIELab color space, and represents the image on three channels, namely, lightness, chroma, and hue which enables direct access to the chrominance and hue channels, as shown in Figure 4.

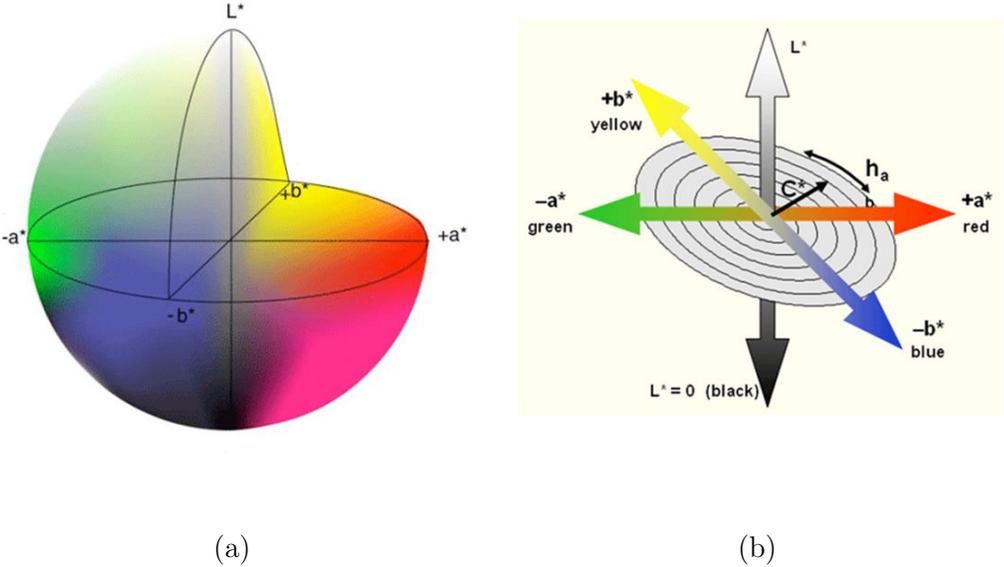

(a) (b)

Fig. 4. a) CIELab color space. b) CIELCh color space

After converting the original image to the CIELCh color space, the histogram along the hue component is calculated and sorted in descending order. The accumulated histogram is counted from the most numerous color points to the least, until it reaches 99% of the whole histogram, as shown in Figure 5. In this way, the number of primary colors appearing in the image can be estimated. This number will be used in the color-reduced representation that will be discussed below. The question of the number of bins will be addressed in the experimental section. In short, according to our experiments, 40 colors are generally sufficient for a normal document image without losing substantial color data.

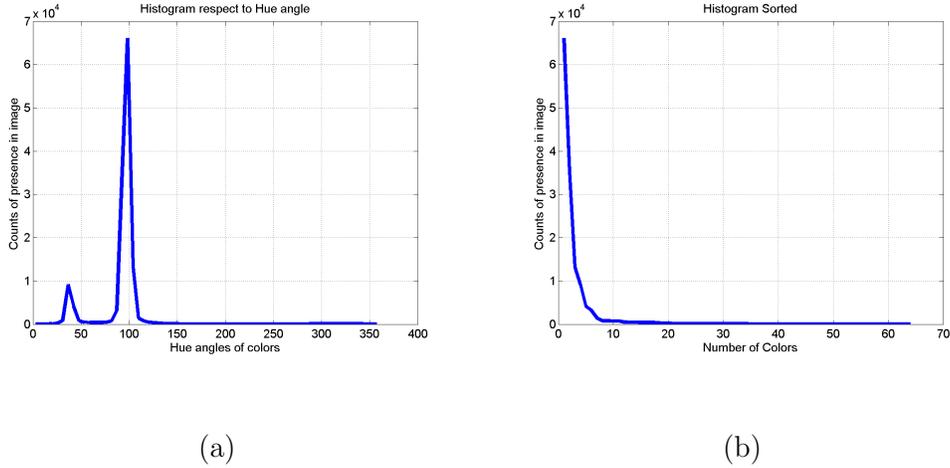

(a)                  (b)

Fig. 5. a) Histogram of hue component of the original image. The horizontal axis is the hue angle, ranging from 0 to 359 degrees, and the vertical axis is the distribution of the color points. b) The sorted histogram. The accumulated histogram is cut at 99% of the whole histogram, where it gives the number of primary colors. (The corresponding test image is shown in Figure 2(a)).

Apparently, the number of primary colors, calculated by the proposed histogram method, corresponds to the number of different hue angles presented in the image, not to the number of different colors (which may be located in the same hue angle but with a different degree of saturation). However, as our study is restricted to the special case of document images in which most colors are rather different in hue angle (such as black or blue for the main text ink with red for the marginal text ink) than in saturation variation, the assumption that the number of primary hue angles can replace the overall primary color number is still valid.

Thereafter, an indexed version of a color image, as discussed in many classical image processing books such as [14], is constructed with the color number obtained by the above-mentioned histogram thresholding.

The reduction in the number of colors can be seen from the data distribution in the RGB space (see Figure 6). Although the data volume is considerably reduced, it still represents the essential part of the original data cloud. It is worth noting that the number of colors is designed to be image-content dependent (99% of the hue histogram) and will be higher for images rich in color and lower otherwise.

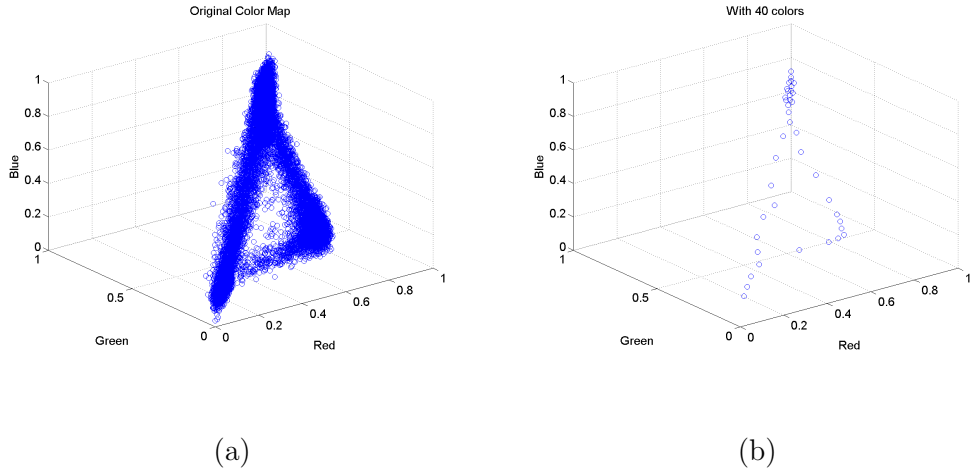

(a)          (b)

Fig. 6. a) The original image color distribution in the RGB color space of Figure 2(a). b) Color-reduced version with 40 colors.

5.2  *Color interpolation after color reduction*

Applying color interpolation after color reduction seems to be contradictory. However, this substep is necessary to preserve the smoothness of the color-reduced image. The color reduction process aims to reduce all color points to primary ones in order to generate a clearer target for binarization. However, this process can introduce some color variations which might lead to bigger gaps between the primary colors than before color reduction. Indeed, the major color differences are still kept between these primary colors (and also, the appearance of the image is similar to that of its original version), whereas the variation in colors is not necessarily reduced, as illustrated in Figure 7.

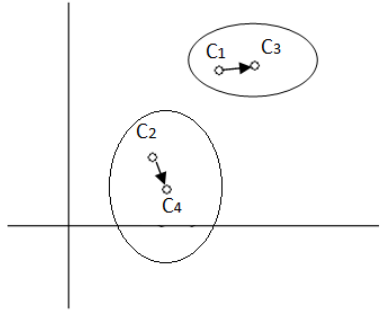

Fig. 7. An example of an increase in color variance after the color reduction step. C1 (the original color) is mapped to C3 (the closest primary color), and C2 to C4. The color difference between C1 and C2 increases with the distance between their new representers, C3 and C4.

Since a low-variance version of the input image is desired for binarization, an interpolation substep is applied after color reduction. At first, primary color points are classified into three classes, as shown in Figure 8(a). The first class comprises the color points that are close enough to each other (for example, zone A), whereas those color points that are too far away from each other (zone C) are classified in the second class. Both these classes are excluded from interpolation to avoid over-interpolated data. The rest of the points are grouped into the third class (for example, zone B). The color points in this group are interpolated based on the distances between them. In this work, the Euclidean distance in the RGB color space is used. It is worth noting that although the perceptual uniformity of color difference can be preserved better using color spaces such as CIECAM97s, CIECAM02, iCAM, or IPT because of their better hue consistency, a reduced numeric variance of color distribution is desired for further processing in this work, and an interpolation substep is sufficient enough because we are not looking for a perception-based hue constancy. Therefore, the fast computable Euclidean distance in the RGB or CIELAB color spaces is sufficient. To control the interpolation process, a minimum and a maximum threshold value are used. In order to make the threshold values adaptive to the input image, they are designed to be a function of the standard deviation of the input image. Based on our tests, the following thresholds give good performance at the interpolation substep:

$$d_{min} = 0.3\sigma_{max},$$
$$d_{max} = 2\sigma_{max},$$
$$\sigma_{max} = \max_{i=1,\cdots,n_c} \sigma_{u_i}$$

where $d_{min}$ and $d_{max}$ are the minimum and maximum thresholds, and $\sigma_{max}$ is the maximum standard deviation of the RGB channels in the original image. The gaps between points within the two thresholds ($d_{min}$ and $d_{max}$) are bridged after interpolation. The interpolated color data are also shown in Figure 8(b).

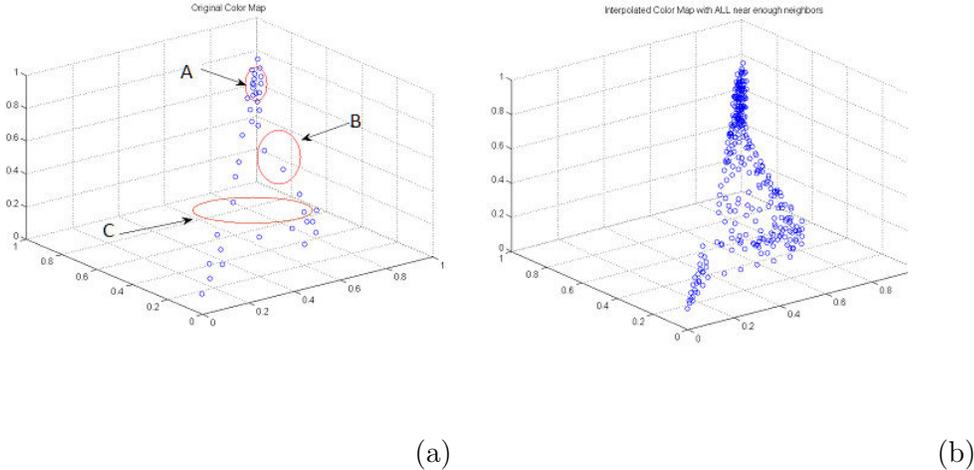

(a)        (b)

Fig. 8. a) Primary colors (by color-reduction). b) Color points after interpolation.

Figure 9 shows the output of the color reduction with the interpolation step.

As can be seen, although the number of colors has been reduced dramatically (from thousands to 40), the major color appearance is still maintained because that 99% of the histogram has been kept. In the experimental section, it will be shown that a number of 40 reduced colors is optimal for the highest binarization performance. The color-interpolated output contains fewer noise-like colors and there is less variation in the color, which makes the subsequent binarization processes more robust. The binarized versions are also shown in the figure, For more details, please see the experimental section. Although it can be argued that most of this noise can be corrected using morphological operators, preventing them from the source increases the generalizability of the model and also reduces the computational complexity associated with the corrections steps.

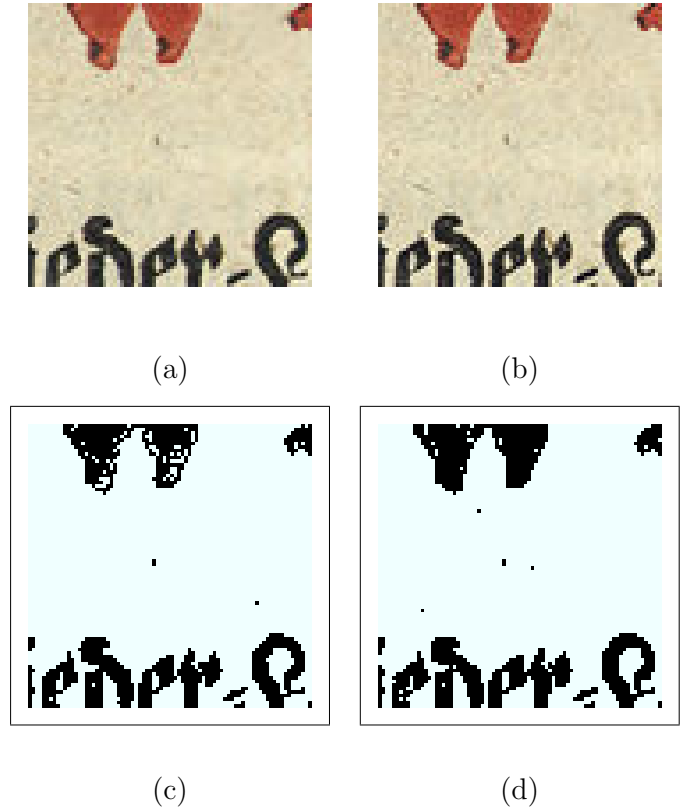

Fig. 9. a) The original image. b) The color-reduced and -interpolated image (40 primary colors plus interpolated colors). c) and d) The binarized versions of (a) and (b). As can be seen, the color reduction and interpolation step avoids addition of noise to the output.

It is worth noting that the color reduction and interpolation step can be seen as a clustering step in which the pixels are grouped in many clusters which are at a specific distance from each other (something between $d_{min}$ and $d_{max}$). However, it may require special modeling in order to handle the associated computational cost because of large number of sample (presence of millions of pixels on a typical image) which is beyond the scope of this work. The proposed color reduction and interpolation step provides sufficient and fast solution as will be seen in the experimental section.

## 6 Experimental Results and Discussion

*6.1 Subjective evaluation*

The images used in the experiments are taken from several datasets of real normal or old document images. The resolution, stroke size, illumination contrast, background complexity and noise levels are variable from one image to another. The datasets are: i) one image from the DIBCO'09 contest dataset [12], ii) 22 images from the ICDAR'03 Robust Reading dataset, iii) 1172 scene text images in English or Mixed of English and Korean from the KAIST dataset [17, 16] [3], and iv) two old manuscripts, courtesy of the Juma Al Majid Center for Cultural Heritage (Dubai) [4], which together contain about 500 pages. In Figure 10(a), a sample image from dataset (i) is shown which contains text with different colors and different stroke widths. The ground truth images of the ICDAR'03 Robust Reading dataset are manually created by the authors, and will be published publicly on one of dataset hosting servers in near future.

The effectiveness of the color symmetrization used in the dual transform can be seen subjectively from the gray-level images obtained by various methods (shown in Figure 10). For this purpose, we use four methods: 1) the average method, 2) the PCA method, 3) the min-average method [7], and 4) the proposed dual transform method. Based on the visual inspection of the gray-level images obtained, the result of the proposed dual transform method looks more

---

[3] Available on TC-11 website: `http://www.iapr-tc11.org/mediawiki/index.php/KAIST_Scene_Text_Database`

[4] http://www.almajidcenter.org/English/Pages/default.aspx

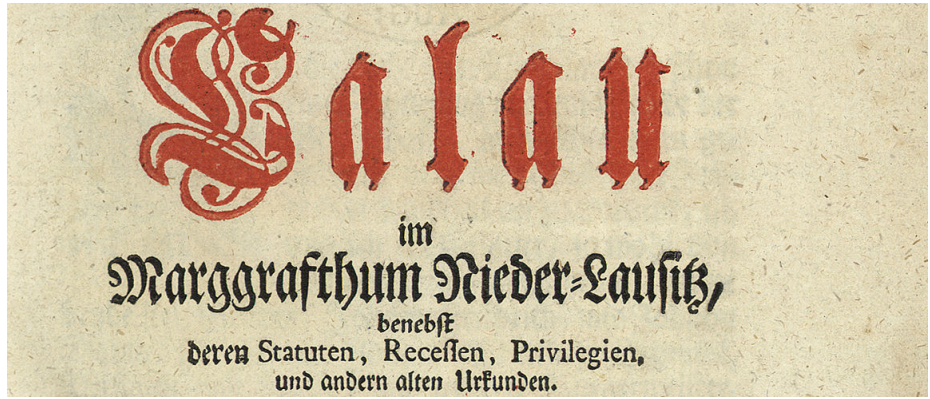
(a)

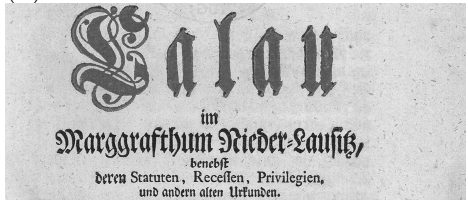
(b)

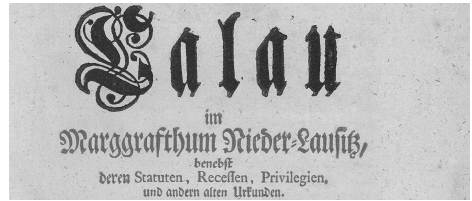
(c)

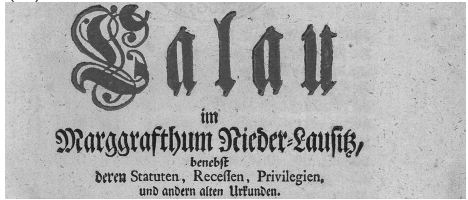
(d)

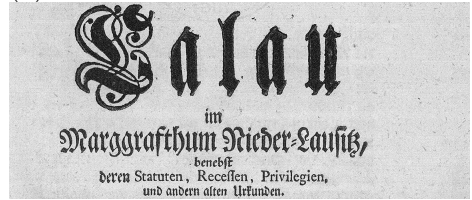
(e)

Fig. 10. Illustration of the visual performance of different color to gray conversion methods. a) The original color image. b) The output of the average method. c) The output of the PCA method. d) The output of the min-average method. e) The output of the proposed method.

uniform than that of the other methods. This is because it can produce low-variable intensity text regions. In other words, this method ensures the quality and preservation of meaningful textual data in the converted gray-level images by reducing their dissimilarity after conversion to gray. Figure 11 provides the profile of the variation on the text regions for all four methods. To calculate the variation on these regions, the text regions are identified using the available ground-truth image. As expected, the dual transform and min-average methods achieve a low level of variation. The low variation on the text regions enables the subsequent binarization methods to extract the text regions more easily.

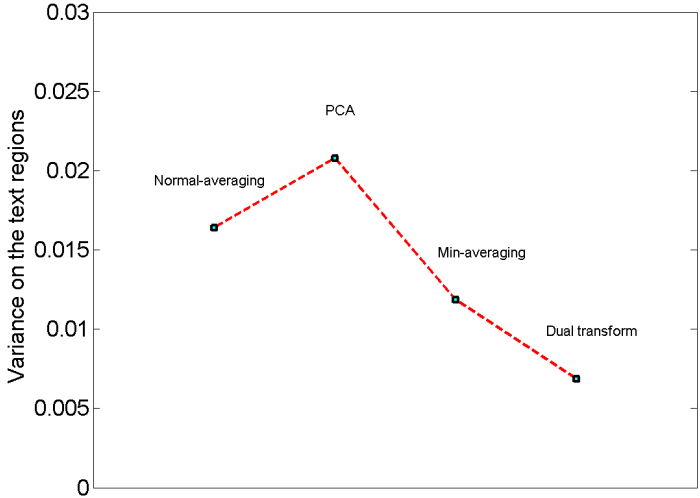

Fig. 11. The profile of the variation on the text regions of gray images of the four conversion methods. The proposed dual transform method achieves the lowest variation, as expected.

Another set of test images is shown in Figure 12. It can be seen that the text has been enhanced in the dual transform outputs, that are shown in the fourth row. Also, the contrast between text and background is increased, despite the fact that there are different colors in the original images. In contrast, the other methods produce images in which two original texts with different colors always remain different in gray. This drawback can lower the performance of the subsequent image binarization methods. The average method, which is based on a simple averaging of color channels, generates non homogeneous intensity on the gray-level image, which is not well suited to the binarization task. The PCA method, which is based on reducing the correlated information and keeping only the non correlated information, shows a big intensity variation between two original texts of different colors. Although this method improves the contrast of the image, it cannot put all the document text on the same gray level. The min-average method performs better than the previous two methods, thanks to the use of the minimum color channel to calculate the gray value. This makes it less sensitive to the input color variations. The output gray values on the text regions are then at similar levels, even if they have different colors in the input color image. As discussed in the Introduction, the min-average method is a non-adaptive method, and therefore it cannot adapt to the particular distribution of color data of each input image. The proposed dual transform method is capable of putting color regions on a similar intensity level, thanks to its affine transformation of the color distribution. With this method, the homogeneity of stroke regions is well preserved because of the consistency of the intensity distribution within the text regions. The proposed method can be seen as a luminance-independent contrast enhancement.

## 6.2  Objective evaluation

For the purposes of objective evaluation, the performance of a binarization method on the gray-level images is used as a measure to evaluate the quality of the gray converters. The trueness of the binarized images is evaluated using the F-measure [32]. Because many of input images from the KAIST dataset contain text-like artifacts which are not considered in the ground-truth images, we limit the calculations of the F-measure to the areas around the ground truth text regions. The radius of the text regions is locally calculated from the ground truth images, and the area with double the local text radius is used in

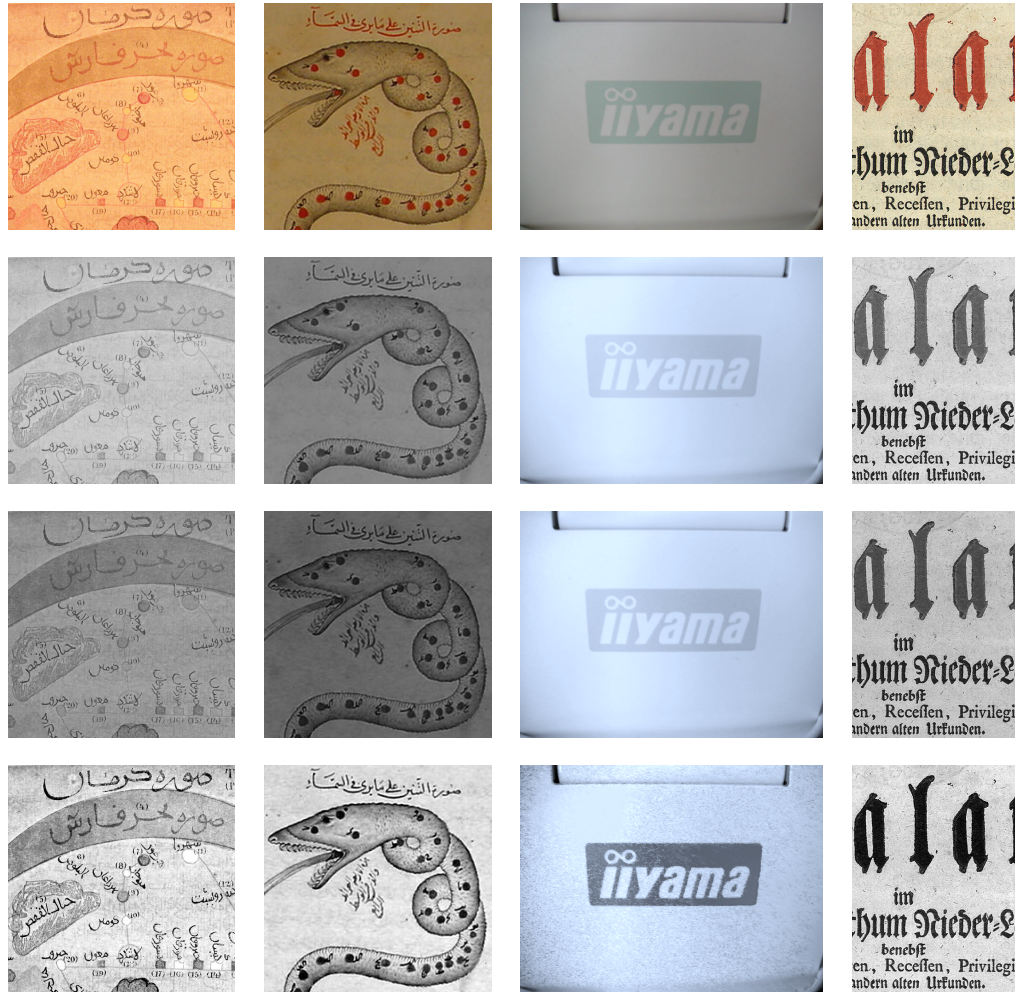

Fig. 12. Another set of images and their corresponding gray-level images obtained using some of the methods in Figure 10. First row: the original images. Second row: the output of the average method. Third row: the output of the min-average method. Fourth row: the output of the dual transform method.
the calculations.

We chose the grid-based Sauvola method [7] as the binarization method, because of its local adaptation feature, which enables it to capture the variations in intensity over the document domain. High-performance methods, such as the grid-based multi-scale Sauvola method [7] or edge-based methods [18], were not selected, in order to better reveal the difference between the gray converters. As we use a single binarization method in all cases, the choice of the binarization method is arbitrary. Because of availability of the source code of the grid-based Sauvola method to the authors, we use this method in the evaluations. The parameters of the grid-based Sauvola method are set as follows: the window size is set to half the average line height [7], which is defined as the distance between two successive lines on the document image, and the value of $k$ is set to 0.2.

As discussed in subsection 3.2.1, a color reduction and interpolation step is used within the dual transform method in order to reduce the variation introduced because of the dual transform method. Here, the efficiency of this step is first evaluated subjectively. The evaluation is performed by comparing the binarization of the output of the dual transform with different numbers of reduced colors. By the number of reduced colors, we mean the number of bins used in the histogram along the hue component, as described in section 5.1. Figure 13 shows the binarization results with the number of colors ranging from 6 (Figure 13(a)) to 40 (Figure 13(c)). Figure 13(d) is obtained with the full-color image (i.e. without any color reduction). Visually, it can be seen that the binarized results for the color-reduced images with the high number of reduced colors (40) are better in terms of regional text homogeneity than the results for the full-color images. The 50 colors and 60 colors used cases are not shown because there is no significant visual difference with the 40 colors used case. It is worth noting that the colors reduction can increase the performance of the dual transform method in the sense of reducing the undesirable variation, which can be considered as noise, added by the method.

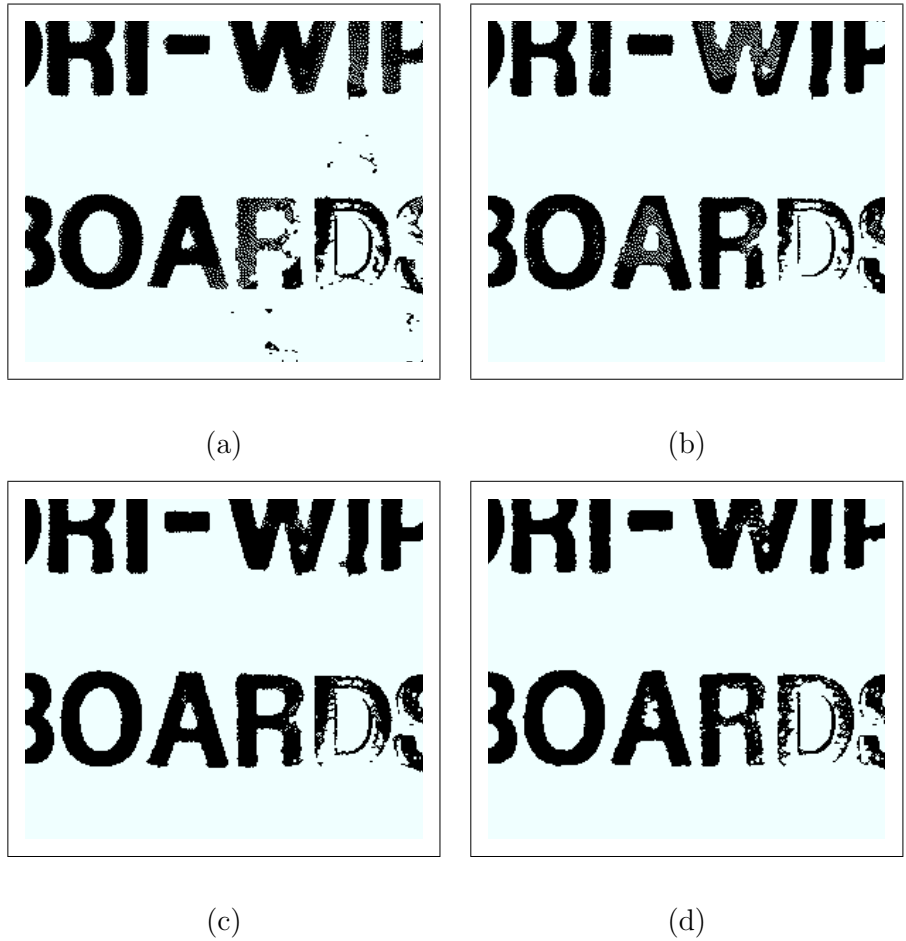

(a)      (b)

(c)      (d)

Fig. 13. The effect of color reduction on the performance of the dual transform method. a) 6 colors are used. b) 12 colors are used. c) 40 colors are used. d) no color reduction. The original image is shown in Figure 2(b).

An objective evaluation of the binarized images in terms of the number of reduced colors has also been carried out, in order to show the effectiveness of color reduction and interpolation. Table 1 shows that with reducing to 40 colors, which corresponds to the primary color number and covers 99% of the hue component histogram (see section 5), a better binarization result is obtained. This means that, in this case, the input gray-level image supplied to the binarization process contains more uniform text and helps in pixel

| Degree of color reduction | Precision | Recall | F-measure |
|---:|:---:|:---:|:---:|
| No reduction | 0.98789 | 0.84527 | 90.81565 |
| Reduction to 6 colors | 0.98624 | 0.79284 | 87.49248 |
| Reduction to 12 colors | 0.98820 | 0.80405 | 88.30734 |
| Reduction to 40 colors | 0.89975 | 0.95080 | **91.49068** |
| Reduction to 50 colors | 0.89423 | 0.95310 | 91.25055 |
| Reduction to 60 colors | 0.88780 | 0.95639 | 91.08012 |

Table 1

The impact of color reduction and interpolation on the performance of the proposed method.

classification. It is worth noting that the precision, recall and F-measure values presented in Table 1 are the average values over the whole set of images, and do not obey the F-measure relation to precision and recall.

In the second part of the evaluation, the binarization performance of the proposed method is compared to the other gray converters. The binarized results of one of the input images are shown in Figure 14. From the figure, we can note that the output of the proposed method (see Figure 14(d)) is better than that of the others. This means that the proposed method is capable of offering a uniform input gray image for the binarization task that is required by most binarization processes. The F-measure evaluation is also performed. Table 2 shows that the application of the proposed dual transform in document image binarization improves the performance at least 10% in terms of the F-measure results. Thus, by using the proposed method, the subsequent

| Method name | Precision | Recall | F-measure |
|---|---|---|---|
| Average method | 0.98103 | 0.68390 | 79.90981 |
| Average method[(a)] | 0.97974 | 0.69818 | 80.83722 |
| PCA method | 0.97988 | 0.67320 | 79.17274 |
| Min-average method | 0.98201 | 0.66580 | 78.66541 |
| Proposed Dual transform method | 0.89975 | 0.95080 | **91.49068** |

Table 2

Performance of the grid-based Sauvola binarization method in terms of Precision, Recall, and the F-measure for the various gray conversion methods on the selected datasets. See the text for more details. [(a)] Without color reduction and interpolation.

binarization method is more efficient and achieved better results.

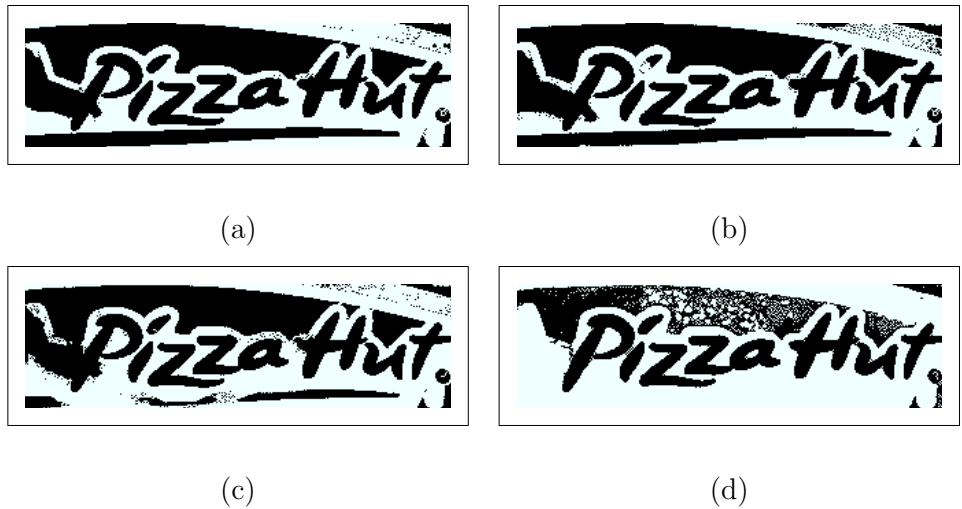

(a)  (b)

(c)  (d)

Fig. 14. The outputs of the grid-based Sauvola binarization method applied to the results of the four conversion methods (on the input image shown in Figure 2(c)): a) The average method. b) The PCA method. c) The min-average method. d) The proposed method. The original image is shown in Figure 2(c).

In terms of the computational complexity, the proposed method is the most complex method compared to the others. However, the total conversion time for an image of size $1100 \times 500$ pixels is less than a second for the current unoptimized implementation that seems to be acceptable.

## 7 Conclusion and Future Prospects

A conversion method, based on dual transform, has been introduced for converting color/multi-spectral images to gray-level images. This method balances the color data distribution in the color space between all color channels using an affine transformation. The balanced distribution of energy among all channels, obtained using the dual transform, makes the resulting gray images less sensitive to the color data of the original images. This helps subsequent processing steps, such as binarization, to achieve better results. The method uses a color reduction and interpolation step in order to control the amplified color variations after the dual transform. The performance of the method has been proven to be promising by using subjective and objective evaluations against various datasets. The method works as a luminance-independent contrast enhancement.

It is worth noting that the dual transform can be considered as an optimized linear transformation for the color-insensitive conversion of color/multi-spectral images. As a prospect for future study, the development of nonlinear image-based transformations, constructed using Riemannian manifolds, will be considered for color-insensitive gray conversion. In this way, the curvilinear structure of each image can be learned and used in the conversion. Another prospect for future study is the application of the proposed color reduction technique to the problem of color image denoising.

**Acknowledgments**

The authors thank the NSERC of Canada for their financial support.